\def\year{2021}\relax
\documentclass[letterpaper]{article} % DO NOT CHANGE THIS
\usepackage{aaai21}  % DO NOT CHANGE THIS
\usepackage{times}  % DO NOT CHANGE THIS
\usepackage{helvet} % DO NOT CHANGE THIS
\usepackage{courier}  % DO NOT CHANGE THIS
\usepackage[hyphens]{url}  % DO NOT CHANGE THIS
\usepackage{graphicx} % DO NOT CHANGE THIS
\usepackage{graphicx}  % DO NOT CHANGE THIS
\usepackage{natbib}  % DO NOT CHANGE THIS AND DO NOT ADD ANY OPTIONS TO IT
\usepackage{caption} % DO NOT CHANGE THIS AND DO NOT ADD ANY OPTIONS TO IT
\usepackage[switch]{lineno}  %
\usepackage{amsthm}
\newtheorem{example}{Example}
\usepackage{booktabs}
\usepackage{subcaption}

\usepackage[linesnumbered,vlined,ruled]{algorithm2e}
\SetKwProg{Fn}{Function}{}{}
\SetArgSty{textnormal} % text style in if condition 
\SetCommentSty{emph}
\usepackage{setspace}

\usepackage{multirow}
\usepackage{multicol}
\usepackage{cleveref}
\usepackage{comment}

\newcommand{\jl}[1]{}{}%{\textcolor{blue}{[JL: #1]}}
\newcommand{\windowed}{RHCR}
\title{Lifelong Multi-Agent Path Finding in Large-Scale Warehouses\protect\thanks{This paper is an extension of \cite{LiAAMAS20b}. }}
\author{
    Jiaoyang Li,\textsuperscript{\rm 1} %\protect\thanks{Jiaoyang Li performed the research during her visit at Monash University.}
    Andrew Tinka,\textsuperscript{\rm 2}
    Scott Kiesel,\textsuperscript{\rm 2} \\
    Joseph W. Durham,\textsuperscript{\rm 2}
    T. K. Satish Kumar,\textsuperscript{\rm 1}
    Sven Koenig\textsuperscript{\rm 1}\\
}
\affiliations{
    \textsuperscript{\rm 1} University of Southern California\\
    \textsuperscript{\rm 2} Amazon Robotics \\
    jiaoyanl@usc.edu,
\{atinka, skkiesel, josepdur\}@amazon.com,
tkskwork@gmail.com,
skoenig@usc.edu 
}

\begin{document}
%\linenumbers  %
\maketitle

\begin{abstract}
Multi-Agent Path Finding (MAPF) is the problem of moving a team of agents to their goal locations without collisions. In this paper, we study the lifelong variant of MAPF, where agents are constantly engaged with new goal locations, such as in large-scale automated warehouses.
We propose a new framework Rolling-Horizon Collision Resolution (\windowed) for solving lifelong MAPF by decomposing the problem into a sequence of Windowed MAPF instances, where a Windowed MAPF solver resolves collisions among the paths of the agents only within a bounded time horizon and ignores collisions beyond it.
\windowed{} is particularly well suited to generating pliable plans that adapt to continually arriving new goal locations.
We empirically evaluate \windowed{} with a variety of MAPF solvers and show that it can produce high-quality solutions for up to 1,000 agents (= 38.9\% of the empty cells on the map) for simulated warehouse instances, significantly outperforming existing work.
\end{abstract}

\sloppy
\section{Introduction}

\emph{Multi-Agent Path Finding} (MAPF) is the problem of moving a team of agents %on a  graph
from their start locations to their goal locations while avoiding collisions. %~\cite{SternSoCS19}.
The quality of a solution is measured by \emph{flowtime} (the sum of the arrival times of all agents at
their goal locations) or \emph{makespan} (the maximum of the arrival times of all agents at their goal locations). MAPF is NP-hard to solve optimally~\cite{YuAAAI13}.
%Although MAPF is NP-hard to solve optimally~\cite{YuAAAI13}, many MAPF algorithms have been developed in recent years, including rule-based algorithms~\cite{LunaIJCAI11,WildeAAMAS13}, prioritized planning~\cite{MaAAAI19,OkumuraIJCAI19}, reduction-based algorithms~\cite{LamIJCAI19,SurynekIJCAI19}, A*-based algorithms~\cite{GoldenbergAIJ14,Wagner15} and dedicated search-based algorithms~\cite{SharonAIJ13,BarrerSoCS14,LiIJCAI19}.

MAPF has numerous real-world applications, such as
autonomous aircraft-towing vehicles~\cite{MorrisAAAI16}, %office robots~\cite{VelosoIJCAI15}, 
video game characters~\cite{LiAAMAS20a}, and quadrotor swarms~\cite{HoenigTRO18}. %, and warehouse robots~\cite{LiuAAMAS19}.
%In the near future, for instance, aircraft-towing vehicles might navigate autonomously to aircraft and tow them from the runways to their gates.
Today, in automated warehouses, mobile robots called \emph{drive units} already autonomously move inventory pods or flat packages from one location to another~\cite{WurmanAAAI07,KouAAAI20}.
However, MAPF is only the ``one-shot'' variant of the actual problem
in many applications.
Typically, after an agent reaches its goal location, it does not stop and wait there forever. Instead, it is assigned a new goal location and required to keep moving, which is referred to as \emph{lifelong MAPF}~\cite{MaAAMAS17} and characterized by agents constantly being assigned new goal locations.

Existing methods for solving lifelong MAPF include (1) solving it as a whole~\cite{NguyenIJCAI17}, (2) decomposing it into a sequence of MAPF instances where one replans paths at every timestep for all agents~\cite{WanICARCV18,GrenouilleauICAPS19}, and (3) decomposing it into a sequence of MAPF instances where one plans new paths at every timestep for only the agents with new goal locations~\cite{CapICAPS15,MaAAMAS17,LiuAAMAS19}.

In this paper, we propose a new framework \emph{Rolling-Horizon Collision Resolution} (\windowed) for solving lifelong MAPF where we decompose lifelong MAPF into a sequence of Windowed MAPF instances and replan paths once every $h$ timesteps (replanning period $h$ is user-specified) for interleaving planning and execution.
A \emph{Windowed MAPF} instance is different from a regular MAPF instance in the following ways:
\begin{enumerate}
    \item it allows an agent to be assigned a sequence of goal locations within the same Windowed MAPF episode, and
    \item collisions need to be resolved only for the first $w$ timesteps (time horizon $w \geq h$ is user-specified).
\end{enumerate}%
The benefit of this decomposition is two-fold. First, it keeps the agents continually engaged, avoiding idle time, and thus increasing throughput. Second, it generates pliable plans that adapt to continually arriving new goal locations. In fact, resolving collisions in the entire time horizon (i.e., $w=\infty$) is often unnecessary since the paths of the agents can change as new goal locations arrive. 

We evaluate \windowed{} with various MAPF solvers, namely CA*~\cite{SilverAIIDE05} (incomplete and suboptimal),  PBS~\cite{MaAAAI19} (incomplete and suboptimal), ECBS~\cite{BarrerSoCS14} (complete and bounded suboptimal), and CBS~\cite{SharonAIJ15} (complete and optimal). 
We show that, for each MAPF solver, using a bounded time horizon yields similar throughput as using the entire time horizon but with a significantly smaller runtime.
We also show that \windowed{} outperforms existing work and can scale up to 1,000 agents (= 38.9\% of the empty cells on the map) for simulated warehouse instances.

\section{Background}~\label{sec:background}
In this section, we first introduce several state-of-the-art MAPF solvers and then discuss existing research on lifelong MAPF. We finally review the elements of the bounded horizon idea that have guided previous research.

%%%%%%%%%%%%%%%%%%%%%%%%%%%%%%%%%%%%%%%%%%%%%%%
\subsection{Popular MAPF Solvers}\label{sec:algo}

Many MAPF solvers have been proposed in recent years, including rule-based solvers~\cite{LunaIJCAI11,WildeAAMAS13}, prioritized planning~\cite{OkumuraIJCAI19}, compilation-based solvers~\cite{LamIJCAI19,SurynekIJCAI19}, A*-based solvers~\cite{GoldenbergAIJ14,Wagner15}, and dedicated search-based solvers~\cite{SharonAIJ13,BarrerSoCS14}.
We present four representative MAPF solvers.

\paragraph{CBS}
%\noindent\textbf{CBS}\quad
Conflict-Based Search (CBS)~\cite{SharonAIJ15} is a popular two-level MAPF solver that is complete and optimal.
At the high level, CBS starts with a root node that contains a shortest path for each agent (ignoring other agents). It then chooses and resolves a collision by generating two child nodes, each with an additional constraint that prohibits one of the agents involved in the collision from being at the colliding location at the colliding timestep. It then calls its low level to replan the paths of the agents with the new constraints. CBS repeats this procedure
until it finds a node with collision-free paths. 
CBS and its enhanced variants~\cite{GangeICAPS19,LiIJCAI19,LiICAPS20} are among the state-of-the-art optimal MAPF solvers.

\paragraph{ECBS}
%\noindent\textbf{ECBS}\quad
Enhanced CBS (ECBS)~\cite{BarrerSoCS14} is a complete and bounded-suboptimal variant of CBS.
The bounded suboptimality (i.e., the solution cost is a user-specified factor away from the optimal cost) is achieved by using focal search~\cite{PearTPAMIl1982}, instead of best-first search, in both the high- and low-level searches of CBS.
ECBS is the state-of-the-art bounded-suboptimal MAPF solver.

\paragraph{CA*}
%\noindent\textbf{CA*}\quad
Cooperative A* (CA*)~\cite{SilverAIIDE05} is based on a simple prioritized-planning scheme: Each agent is given a unique priority and computes, in priority order, a shortest path 
that does not collide with the (already planned) paths of agents with higher priorities. 
CA*, or prioritized planning in general, is widely used in practice due to its small runtime.
However, it is suboptimal and incomplete since its predefined priority ordering can sometimes result in solutions of bad quality or even fail to find any solutions for solvable MAPF instances.

\paragraph{PBS}
%\noindent\textbf{PBS}\quad
Priority-Based Search (PBS)~\cite{MaAAAI19} combines the ideas of CBS and CA*.
The high level of PBS is similar to CBS except that, when resolving a collision, instead of adding additional constraints to the resulting child nodes, PBS assigns one of the agents involved in the collision a higher priority than the other agent in the child nodes. The low level of PBS is similar to CA* in that it 
plans a shortest path that is consistent with the partial priority ordering generated by the high level.
PBS outperforms many variants of prioritized planning in terms of solution quality but
is still incomplete and suboptimal.

%%%%%%%%%%%%%%%%%%%%%%%%%%%%%%%%%%%%%%%%%%%%%%%%%%
\begin{figure}[t]
    \begin{subfigure}{\columnwidth}
        \centering
        \includegraphics[height=2.9cm]{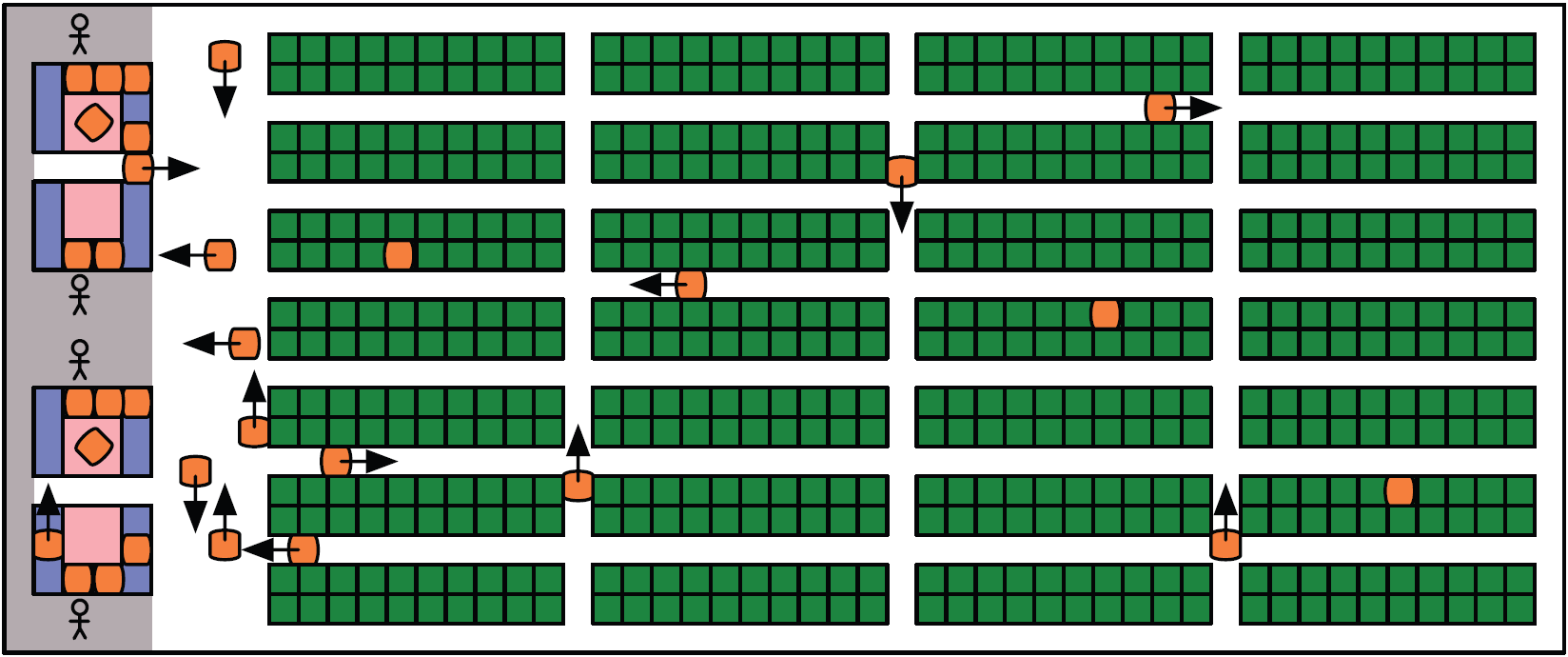} 
        \caption{Fulfillment warehouse map, borrowed from~\protect\cite{WurmanAAAI07}.}\label{fig:kiva}
    \end{subfigure}
    \begin{subfigure}{\columnwidth}
        \centering
        \includegraphics[height=3.5cm]{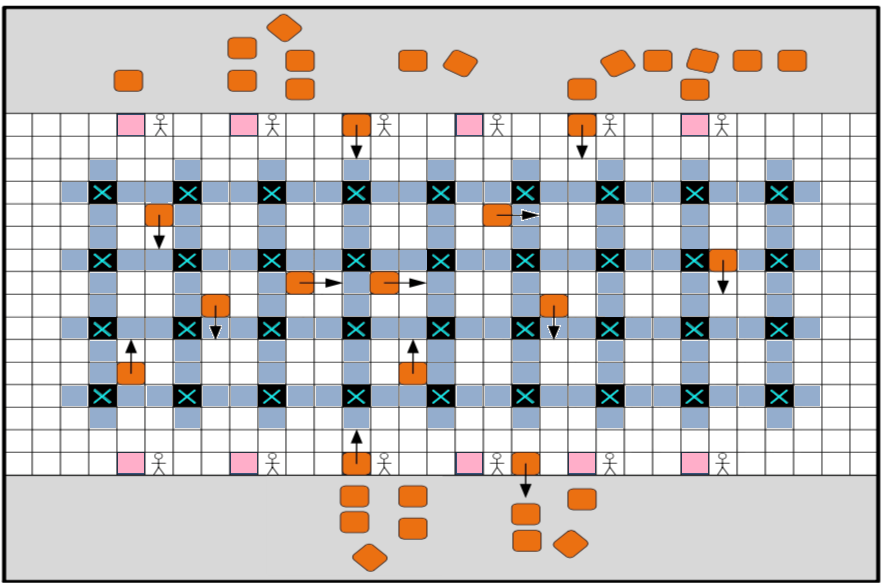} \caption{Sorting center map, modified from~\protect\cite{WanICARCV18}.}\label{fig:distribution-center}
    \end{subfigure}
    \caption{A well-formed fulfillment warehouse map and a non-well-formed sorting center map. Orange squares represent robots. In (a), the endpoints consist of the green cells (representing locations that store inventory pods) and blue cells (representing the work stations). In (b), the endpoints consist of the blue cells (representing locations where the drive units drop off packages) and pink cells (representing the loading stations). Black cells labeled ``X'' represent chutes (obstacles). } \label{fig:amazon}
\end{figure}

\subsection{Prior Work on Lifelong MAPF} \label{sec:prior}

We classify prior work on lifelong MAPF into three categories.

\paragraph{Method (1)}
%\noindent\textbf{Method (1)}\quad
The first method is to solve lifelong MAPF as a whole in an offline setting (i.e., knowing all goal locations a priori) by reducing lifelong MAPF to other well-studied problems.
For example, \citet{NguyenIJCAI17} formulate lifelong MAPF as an answer set programming problem. However, the method only scales up to 20 agents in their paper, each with only about 4 goal locations. This is not surprising because MAPF is a challenging problem and its lifelong variant is even harder.

\paragraph{Method (2)}
%\noindent\textbf{Method (2)}\quad
A second method is to decompose lifelong MAPF into a sequence of MAPF instances where one replans paths at every timestep for all agents. To improve the scalability, researchers have developed incremental search techniques that reuse previous search effort.
For example, \citet{WanICARCV18} propose an incremental variant of CBS that reuses the tree of the previous high-level search. However, it has substantial overhead in constructing a new high-level tree from the previous one and thus does not improve the scalability by much.
\citet{SvancaraAAAI19} use the framework of Independence Detection~\cite{StandleyAAAI10}
to reuse the paths from the previous iteration. It replans paths for only the new agents (in our case, agents with new goal locations) and the agents whose paths are affected by the paths of the new agents. However, when the environment is dense (i.e., contains many agents and many obstacles, which is common for warehouse scenarios),
almost all paths are affected, and thus it still needs to replan paths for most agents.

\paragraph{Method (3)}
%\noindent\textbf{Method (3)}\quad
A third method is similar to the second method but restricts replanning to the paths of the agents that have just reached their goal locations. The new paths need to avoid collisions not only with each other but also with the paths of the other agents.
Hence, this method could degenerate to prioritized planning in case where only one agent reaches its goal location at every timestep. As a result, the general drawbacks of prioritized planning, namely its incompleteness and its potential to generate costly solutions,
resurface in this method.
To address the incompleteness issue, \citet{CapICAPS15} introduce the idea of well-formed infrastructures to enable backtrack-free search. In well-formed infrastructures, all possible goal locations are regarded as endpoints, and, for every pair of endpoints, there exists a path that connects them without traversing any other endpoints.
In real-world applications, some maps, such as the one in \Cref{fig:kiva},  may satisfy this requirement, but other maps, such as the one in \Cref{fig:distribution-center}, may not.
Moreover, 
additional mechanisms are required during path planning.
For example, one needs to force the agents to ``hold'' their goal locations~\cite{MaAAMAS17} or plan ``dummy paths'' for the agents~\cite{LiuAAMAS19} after they reach their goal locations.
Both alternatives result in unnecessarily long paths for agents, decreasing the overall throughput, as shown in our experiments.

\paragraph{Summary}
%\noindent\textbf{Summary}\quad
Method (1) needs to know all goal locations a priori and has limited scalability.
Method (2) can work in an online setting and scales better than Method (1). However, replanning for all agents at every timestep is time-consuming even if one uses incremental search techniques. As a result, its scalability is also limited.
Method (3) scales to substantially more agents than the first two methods, but the map needs to have an additional structure to guarantee completeness. % of the method.
As a result, it works only for specific classes of lifelong MAPF instances.
In addition, Methods (2) and (3) plan at every timestep, which may not be practical since planning is time-consuming.

\subsection{Bounded-Horizon Planing}

Bounded-horizon planning is not a new idea. \citet{SilverAIIDE05} has already applied this idea to regular MAPF with CA*. He refers to it as Windowed Hierarchical Cooperative A* (WHCA*) and empirically shows that WHCA* runs faster as the length of the bounded horizon decreases but also generates longer paths. 
In this paper, we showcase the benefits of applying this idea to lifelong MAPF and other MAPF solvers.
In particular, \windowed{} yields the benefits of lower computational costs for planning with bounded horizons while keeping the agents busy and yet, unlike WHCA* for regular MAPF, decreasing the solution quality only slightly.

\section{Problem Definition} \label{sec:definition}

%Researchers have studied various MAPF models, including those in which agents have different shapes and sizes~\cite{LiAAAI19a}, agents have different velocity constraints~\cite{MaAAAI19a}, and those in which agents face unexpected delays~\cite{MaAAAI17}. A complete survey on MAPF models can be found in~\cite{SternSoCS19}. To the best of our knowledge, our framework can be applied to all these models. However, since we are interested in demonstrating the power of our framework in warehouse applications, we follow the problem definition in~\cite{MaAAMAS17}.

The input is a graph $G=(V,E)$, whose vertices $V$ correspond to locations
and whose edges $E$ correspond to connections between two neighboring locations,
and a set of $m$ agents \{$a_1, \ldots, a_m\}$, each with an initial location.
We study an online setting where we do not know all goal locations a priori. We assume that there is a \emph{task assigner} (outside of our path-planning system) that the agents can request goal locations from during the operation of the system.\footnote{In case there are only a finite number of tasks, after all tasks have been assigned, we assume that the task assigner will assign a dummy task to an agent whose goal location is, e.g., a charging station, an exit, or the current location of the agent.}
Time is discretized into timesteps.
At each timestep, every agent can either \emph{move} to a neighboring location or \emph{wait} at its current location. Both move and wait actions have unit duration.
A \emph{collision} occurs iff two agents
occupy the same location at the same timestep (called a vertex conflict in~\cite{SternSoCS19}) or traverse the same edge in opposite directions at the same timestep (called a swapping conflict in~\cite{SternSoCS19}).
Our task is to plan collision-free paths that move all agents to their goal locations and maximize the \emph{throughput}, i.e., the average number of goal locations visited per timestep. We refer to the set of collision-free paths for all agents as a \emph{MAPF plan}.

%In practice, the task assignment is usually domain-dependent and could have constraints and preferences of its own in different domains.
We study the case where the task assigner is not within our control so that our path-planning system is applicable in different domains. But, for a particular domain, one can design a hierarchical framework that combines a domain-specific task assigner with our domain-independent path-planning system. 
Compared to coupled methods that solve task assignment and path finding jointly, a hierarchical framework is usually a good way to achieve efficiency.
For example, the task assigners in~\cite{MaAAMAS17,LiuAAMAS19} for fulfillment warehouse applications and in~\cite{GrenouilleauICAPS19} for sorting center applications can be directly combined with our path-planning system. We also showcase two simple task assigners, one for each application, in our experiments.

We assume that the drive units can execute any MAPF plan perfectly. Although this seems to be not realistic, there exist some post-processing methods~\cite{HoenigICAPS16} that can take the kinematic constraints of drive units into consideration and convert MAPF plans to executable commands for them that result in robust execution. For example, \citet{HoenigRAL19} propose a framework that interleaves planning and execution and can be directly incorporated with our framework \windowed{}.

\section{Rolling-Horizon Collision Resolution} \label{sec:framework}

Rolling-Horizon Collision Resolution (\windowed{}) has two user-specified parameters, namely the time horizon $w$ and the replanning period $h$. The time horizon $w$ specifies that the Windowed MAPF solver has to resolve collisions within a time horizon of $w$ timesteps.
The replanning period $h$ specifies that the Windowed MAPF solver needs to replan paths once every $h$ timesteps.
The Windowed MAPF solver has to replan paths more frequently than once every $w$ timesteps to avoid collisions, i.e., $w$ should be larger than or equal to $h$.

In every Windowed MAPF episode, say, starting at timestep $t$,
\windowed{} first updates the start location $s_i$ and the goal location sequence $\mathbf{g_i}$ for each agent $a_i$.
\windowed{} sets the start location $s_i$ of agent $a_i$ to its location at timestep $t$. % 
Then, \windowed{} calculates a lower bound on the number of timesteps $d$ that agent $a_i$ needs to visit all remaining locations in $\mathbf{g_i}$, i.e.,
\begin{equation}
    d=\textnormal{dist}(\mathit{s_i},\mathbf{g_i}[0]) + \sum_{j=1}^{|\mathbf{g_i}|-1}\textnormal{dist}(\mathbf{g_i}[j-1], \mathbf{g_i}[j]),
\end{equation}
where dist($x$, $y$) is the distance from location $x$ to location $y$ and $|\mathbf{x}|$ is the cardinality in sequence $\mathbf{x}$.\footnote{Computing $d$ relies on the distance function dist($x$, $y$). Here, and in any other place where dist($x$, $y$) is required, prepossessing techniques can be used to increase efficiency. In particular, large warehouses have a candidate set of goal locations as the only possible values for $y$, enabling the pre-computation and caching of shortest-path trees. }
$d$ being smaller than $h$ indicates that agent $a_i$ might finish visiting all its goal locations and then being idle before the next Windowed MAPF episode starts at timestep $t+h$. To avoid this situation, \windowed{} continually assigns new goal locations to agent $a_i$ until $d \geq h$. 
Once the start locations and the goal location sequences for all agents require no more updates, \windowed{} calls a Windowed MAPF solver to find paths for all agents that move them from their start locations to all their goal locations in the order given by their goal location sequences and are collision-free for the first $w$ timesteps.
Finally, it moves the agents for $h$ timesteps along the generated paths and remove the visited goal locations from their goal location sequences.

\windowed{} uses flowtime as the objective of the Windowed MAPF solver, which is known to be a reasonable objective for lifelong MAPF~\cite{SvancaraAAAI19}.
Compared to regular MAPF solvers, Windowed MAPF solvers need to be changed in two aspects:
\begin{enumerate}
    \item each path needs to visit a sequence of goal locations, and
    \item the paths need to be collision-free for only the first $w$ timesteps.
\end{enumerate}%
We describe these changes in detail in the following two subsections.

\begin{algorithm}[t]
\caption{The low-level search for Windowed MAPF solvers generalizing Multi-Label A*~\protect\cite{GrenouilleauICAPS19}.
%Function dist($x$, $y$) returns the distance from location $x$ to location $y$.
%The difference to location-time A* is shown in blue (i.e., Lines~\ref{line:label},  \ref{line:1}-\ref{line:goal-test}, and \ref{line:h-function}-\ref{line:h}).
} \label{alg:astar}
\KwIn{Start location $s_i$, goal location sequence $\mathbf{g_i}$.}
\BlankLine
$R.\mathit{location} \leftarrow s_i$, $R.\mathit{time} \leftarrow 0$, $R.g \leftarrow 0$\;\label{line:root}
$R.\mathit{label} \leftarrow 0$\;\label{line:label}
$R.h$ $\leftarrow$ \textsc{ComputeHValue}($R.\mathit{location}$, $R.\mathit{label}$)\; \label{line:root-h}
$\mathit{open}$.push($R$)\;\label{line:push-root}
\While{$\mathit{open}$ is not empty \label{line:while}} {
    $P \leftarrow \mathit{open}$.pop()\tcp*{Pop the node with the minimum $f$.} \label{line:pop}
    \If(\tcp*[f]{Update label.}){$P.\mathit{location} = \mathbf{g_i}[P.\mathit{label}]$\label{line:1}} {
        $P.\mathit{label} \leftarrow P.\mathit{label} + 1$\;\label{line:2}
    }
    \If(\tcp*[f]{Goal test.}){$P.\mathit{label} = |\mathbf{g_i}|$\label{line:goal-test}}{
        \Return the path retrieved from $P$\;\label{line:return}
    }
    \ForEach(\tcp*[f]{Generate child nodes.}){child node $Q$ of $P$ \label{line:child1}} {
        $\mathit{open}$.push($Q$)\; \label{line:child2}
    }
}
\Return ``No Solution''\;
\BlankLine

\SetKwProg{Fn}{Function}{:}{}
    \Fn{\textsc{ComputeHValue}(\textbf{Location} $x$, \textbf{Label} $l$) \label{line:h-function}}{
        \Return $\textnormal{dist}(\mathit{x},\mathbf{g_i}[{\mathit{l}]}) + \sum_{j=\mathit{l}+1}^{|\mathbf{g_i}|-1}\textnormal{dist}(\mathbf{g_i}[j-1], \mathbf{g_i}[j])$\; \label{line:h}
    }
\end{algorithm}

%%%%%%%%%%%%%%%%%%%%%%%%%%%%%%%%%%%%%%%%%%%%%%%%%%%%%%%%%%%%%%%%
\subsection{A* for a Goal Location Sequence} \label{sec:mla}

The low-level searches of all MAPF solvers discussed in \Cref{sec:algo} need to find a path for an agent from its start location to its goal location while satisfying given spatio-temporal constraints that prohibit the agent from being at certain locations at certain timesteps.
Therefore, they often use location-time A*~\cite{SilverAIIDE05} (i.e., A* that searches in the location-time space where each state is a pair of location and timestep) or any of its variants. 
However, a characteristic feature of a Windowed MAPF solver is that it plans a path for each agent that visits a sequence of goal locations. Despite this difference, techniques used in the low-level search of regular MAPF solvers can be adapted to the low-level search of Windowed MAPF solvers.
In fact, \citet{GrenouilleauICAPS19} perform a truncated version of this adaptation for the pickup and delivery problem.
They propose Multi-Label A* that can find a path for a single agent that visits two ordered goal locations, namely its assigned pickup location and its goal location.
In \Cref{alg:astar}, we generalize Multi-Label A* to a sequence of goal locations.\footnote{Planning a path for an agent to visit a sequence of goal locations is not straightforward. While one can call a sequence of location-time A* to plan a shortest path between every two consecutive goal locations and concatenate the resulting paths, the overall path is not necessarily the shortest because the presence of spatio-temporal constraints introduces spatio-temporal dependencies among the path segments between different goal locations, e.g., arriving at the first goal location at the earliest timestep may result in a longer overall path than arriving there later. We therefore need \Cref{alg:astar}. }

\Cref{alg:astar} uses the structure of location-time A*. %~\cite{SilverAIIDE05}.
For each node $N$, %we define $N.\mathit{location}$, $N.\mathit{timestep}$, $N.g$, $N.h$ and $N.\mathit{label}$ as the node’s location, timestep, $g$-value, $h$-value, and label, respectively. %The $f$-value of the node is equal to its $g$-value plus its $h$-value. %The state, in our problem, consists of location, orientation and timestep.
we add an additional attribute $N.\mathit{label}$ that indicates the number of goal locations in the goal location sequence $\mathbf{g_i}$ that the path from the root node to node $N$ has already visited. For example, $N.label=2$ indicates that the path has already visited goal locations $\mathbf{g_i}[0]$ and $\mathbf{g_i}[1]$ but not goal location $\mathbf{g_i}[2]$.
\Cref{alg:astar} computes the $h$-value of a node as the distance from the location of the node to the next goal location plus the sum of the distances between consecutive future goal locations in the goal location sequence [Lines~\ref{line:h-function}-\ref{line:h}]. 
In the main procedure, \Cref{alg:astar} first creates the root node $R$ with label 0 and pushes it into the prioritized queue $\mathit{open}$ [Lines~\ref{line:root}-\ref{line:push-root}].
While $\mathit{open}$ is
not empty [Line~\ref{line:while}], the node $P$ with the smallest $f$-value is selected for expansion [Line~\ref{line:pop}]. % in that iteration
If $P$ has reached its current goal location [Line~\ref{line:1}], $P.\mathit{label}$ is incremented [Line~\ref{line:2}].
If $P.\mathit{label}$ equals the cardinality of the goal location sequence [Line~\ref{line:goal-test}], \Cref{alg:astar} terminates and returns the path [Line~\ref{line:return}].
Otherwise, it generates child nodes that respect the given spatio-temporal constraints [Lines~\ref{line:child1}-\ref{line:child2}].
The labels of the child nodes equal $P.\mathit{label}$. Checking the priority queue for duplicates requires a comparison of labels in addition to other attributes.

\begin{figure*}[t]
\centering
\begin{subfigure}[t]{0.39\textwidth}
    \includegraphics[height=3.8cm]{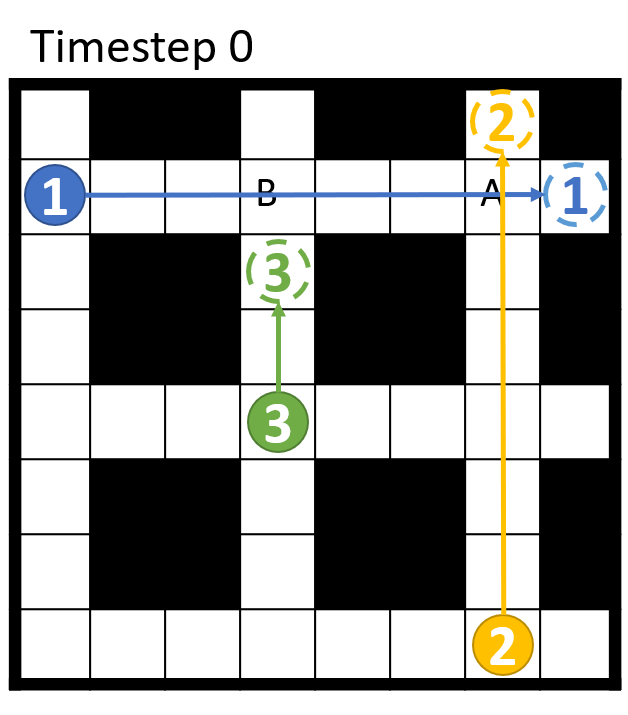} 
    \includegraphics[height=3.8cm]{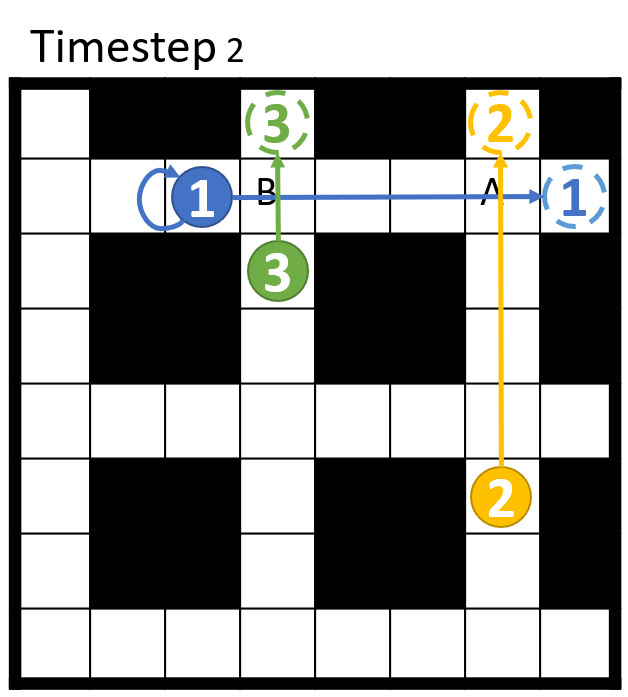}
    \caption{A lifelong MAPF instance with time horizon $w=4$. Agent $a_3$ reaches its goal location at timestep 2 and is then assigned a new goal location.}\label{fig:example3}
\end{subfigure} 
\quad
\begin{subfigure}[t]{0.39\textwidth}
    \includegraphics[height=3.8cm]{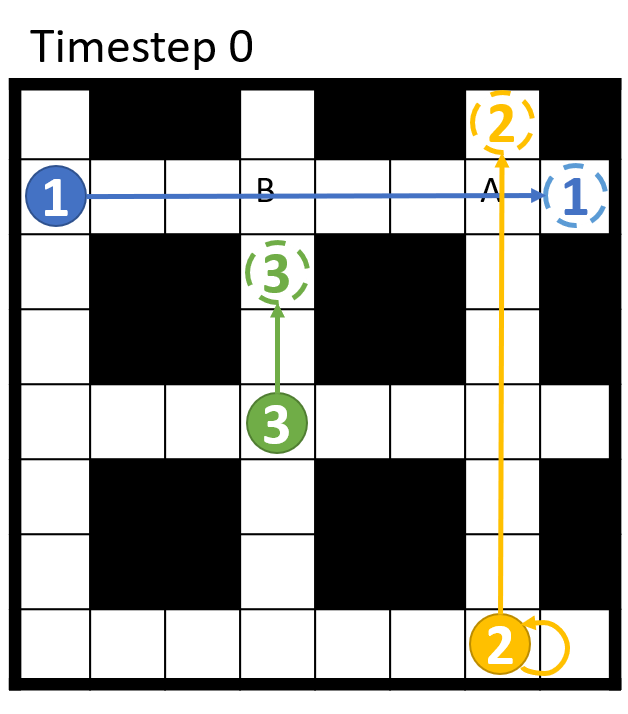} 
    \includegraphics[height=3.8cm]{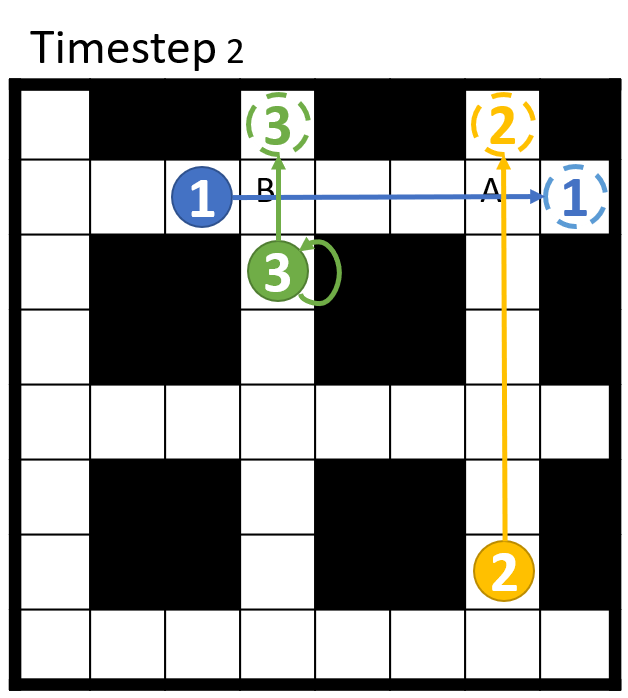}
    \caption{The same lifelong MAPF instance as shown in (a) with time horizon $w=8$.}\label{fig:example5}
\end{subfigure}
\quad
\begin{subfigure}[t]{0.15\textwidth}
    \centering
    \includegraphics[height=2.4cm]{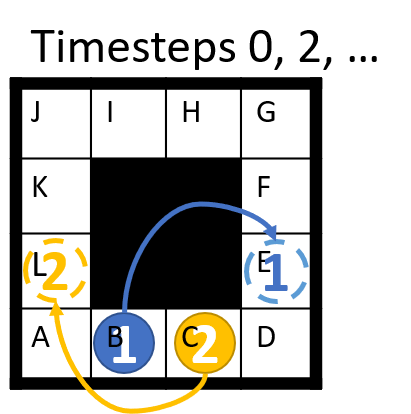} 
    \caption{A lifelong MAPF instance with time horizon $w=2$.}\label{fig:deadlock}
\end{subfigure}
\caption{Lifelong MAPF instances with replanning period $h=2$. Solid (dashed) circles represent the current (goal) locations of the agents.}
\end{figure*}

%%%%%%%%%%%%%%%%%%%%%%%%%%%%%%%%%%%%%%%%%%%%%%%%%%%%%%%%%%%%%%%%
\subsection{Bounded-Horizon MAPF Solvers} \label{sec:window}

Another characteristic feature of Windowed MAPF solvers is the use of a bounded horizon.
Regular MAPF solvers can be easily adapted to resolving collisions for only the first $w$ timesteps. Beyond the first $w$ timesteps, the solvers ignore collisions among agents and assume that each agent follows its shortest path to visit all its goal locations, which ensures that the agents head in the correct directions in most cases.
We now provide details on how to modify the various MAPF solvers discussed in \Cref{sec:algo}.

\paragraph{Bounded-Horizon (E)CBS}
%\noindent\textbf{Bounded-Horizon (E)CBS}\quad
Both CBS and ECBS search by detecting and resolving collisions. In their bounded-horizon variants, we only need to modify the collision detection function. While (E)CBS finds collisions among all paths and can then resolve any one of them, bounded-horizon (E)CBS only finds collisions among all paths that occur in the first $w$ timesteps and can then resolve any one of them. The remaining parts of (E)CBS stay the same. Since bounded-horizon (E)CBS needs to resolve fewer collisions, it generates a smaller high-level tree and thus runs faster than standard (E)CBS.

\paragraph{Bounded-Horizon CA*}
%\noindent\textbf{Bounded-Horizon CA*}\quad
CA* searches based on priorities, where an agent avoids collisions with all higher-priority agents. In its bounded-horizon variant, an agent is required to avoid collisions with all higher-priority agents but only during the first $w$ timesteps. Therefore, when running location-time A* for each agent, we only consider the spatio-temporal constraints during the first $w$ timesteps induced by the paths of higher-priority agents. The remaining parts of CA* stay the same.
Since bounded-horizon CA* has fewer spatio-temporal constraints, it runs faster and is less likely to fail to find solutions than CA*.
Bounded-horizon CA* is identical to WHCA* in~\cite{SilverAIIDE05}.

\paragraph{Bounded-Horizon PBS}
%\noindent\textbf{Bounded-Horizon PBS}\quad
The high-level search of PBS is similar to that of CBS and is based on resolving collisions, while the low-level search of PBS is similar to that of CA* and plans paths that are consistent with the partial priority ordering generated by the high-level search.
Hence, we need to modify the collision detection function of the high level of PBS (just like how we modify CBS) and incorporate the limited consideration of spatio-temporal constraints into its low level (just like how we modify CA*). As a result, bounded-horizon PBS generates smaller high-level trees and runs faster in its low level than standard PBS.

%%%%%%%%%%%%%%%%%%%%%%%%%%%%%%%%%%%%%%%%%%%%%%%%%%%%%%%%
\subsection{Behavior of \windowed{}}

We first show that resolving collisions for a longer time horizon in lifelong MAPF does not necessarily result in better solutions. Below is such an example.

\begin{example} \label{ex:1}
Consider the lifelong MAPF instance shown in \Cref{fig:example3} with time horizon $w=4$ and replanning period $h=2$, and assume that we use an optimal Windowed MAPF solver. At timestep 0 (left figure), all agents follow their shortest paths as no collisions will occur during the first 4 timesteps. Then, agent $a_3$ reaches its goal location at timestep 2 and is assigned a new goal location (right figure). If agents $a_1$ and $a_3$ both follow their shortest paths, the Windowed MAPF solver finds a collision between them at cell B at timestep 3 and forces agent $a_1$ to wait for one timestep.  The resulting number of wait actions is 1. However, if we solve this example with time horizon $w=8$, as shown in \Cref{fig:example5}, we could generate paths with more wait actions. At timestep 0 (left figure), the Windowed MAPF solver finds a collision between agents $a_1$ and $a_2$ at cell A at timestep 6 and thus forces agent $a_2$ to wait for one timestep. Then, at timestep 2 (right figure), the Windowed MAPF solver finds a collision between agents $a_1$ and $a_3$ at cell B at timestep 3 and forces agent $a_3$ to wait for one timestep. The resulting number of wait actions is 2.
\end{example}

Similar cases are also found in our experiments: sometimes \windowed{} with smaller time horizons achieves higher throughput than with larger time horizons.
All of these cases support our claim that, in lifelong MAPF, resolving all collisions in the entire  time horizon is unnecessary, which is different from regular MAPF.
Nevertheless, the bounded-horizon method also has a drawback since using too small a value for the time horizon may generate \emph{deadlocks} that prevent agents from reaching their goal locations, as shown in~\Cref{ex:deadlock}.

\begin{example} \label{ex:deadlock}
Consider the lifelong MAPF instance shown in \Cref{fig:deadlock} with time horizon $w=2$ and replanning period $h=2$, and assume that we use an optimal Windowed MAPF solver.
At timestep 0, the Windowed MAPF solver returns path [B, B, B, C, D, E] (of length 5) for agent $a_1$ and path [C, C, C, B, A, L] (of length 5) for agent $a_2$, which are collision-free for the first 2 timesteps. It does not return the collision-free paths where one of the agents uses the upper corridor, nor the collision-free paths where one of the agents leaves the lower corridor first (to let the other agent reach its goal location) and then re-enters it, because the resulting flowtime is larger than $5+5=10$.
Therefore, at timestep 2, both agents are still waiting at cells B and C. The Windowed MAPF solver then finds the same paths for both agents again and forces them to wait for two more timesteps. Overall, the agents wait at cells B and C forever and never reach their goal locations.
\end{example}

\subsection{Avoiding Deadlocks} \label{sec:completeness}

To address the deadlock issue shown in Example~\ref{ex:deadlock}, we can design a potential function to evaluate the progress of the agents and increase the time horizon if the agents do not make sufficient progress. For example, after the Windowed MAPF solver returns a set of paths, we compute the potential function $P(w)=|\{a_i|\textsc{ComputeHValue}(x_i, l_i) < \textsc{ComputeHValue}(s_i, 0), 1 \leq i \leq m\}|$, where function \textsc{ComputeHValue}($\cdot, \cdot$) is defined on Lines~\ref{line:h-function}-\ref{line:h} in \Cref{alg:astar}, $x_i$ is the location of agent $a_i$ at timestep $w$, $l_i$ is the number of goal locations that it has visited during the first $w$ timesteps, and $s_i$ is its location at timestep 0. 
$P(w)$ estimates the number of agents that need fewer timesteps to visit all their goal locations from timestep $w$ on than from timestep 0 on.
We increase $w$ and continue running the Windowed MAPF solver until $P(w) \geq p$, where $p \in [0, m]$ is a user-specified parameter. This ensures that at least $p$ agents have visited (some of) their goal locations or got closer to their next goal locations during the first $w$ timesteps.

\begin{example}
    Consider again the lifelong MAPF instance in \Cref{fig:deadlock}. Assume that $p=1$. When the time horizon $w=2$, as discussed in \Cref{ex:deadlock}, both agents keep staying at their start locations, and thus $P(2)=0$. When we increase $w$ to 3, the Windowed MAPF solver finds the paths [B, C, D, E] (of length 3) for agent $a_1$ and [C, D, E, ..., K, L] (of length 9) for agent $a_2$. Now, $P(3)=1$ because agent $a_1$ is at cell E at timestep $3$ and needs 0 more timesteps to visit its goal locations. Since $P(3)=p$, the Windowed MAPF solver returns this set of paths and avoids the deadlock.
\end{example}

There are several methods for designing such potential functions, e.g., the number of goal locations that have been reached before timestep $w$ or the sum of timesteps that all agents need to visit their goal locations from timestep $w$ on minus that the sum of timesteps that all agents need to visit their goal locations from timestep $0$ on. In our experiments, we use only the one described above. We intend to design more potential functions and compare their effectiveness in the future.

Unfortunately, \windowed{} with the deadlock avoidance mechanism is still incomplete. %In fact, it is hard to design a complete algorithm for lifelong MAPF, maximizing the throughput while not knowing goal locations in advance. 
Imagine an intersection where many agents are moving horizontally but only one agent wants to move vertically. If we always let the horizontal agents move and the vertical agent wait, we maximize the throughput but lose completeness (as the vertical agent can never reach its goal location). But if we let the vertical agent move and the horizontal agents wait, we might guarantee completeness but will achieve a lower throughput. This issue can occur even if we use time horizon $w=\infty$. Since throughput and completeness can compete with each other, we choose to focus on throughput instead of completeness in this paper. 

\begin{figure}[t]
    \begin{subfigure}{\columnwidth}
        \centering
        \includegraphics[height=3.3cm]{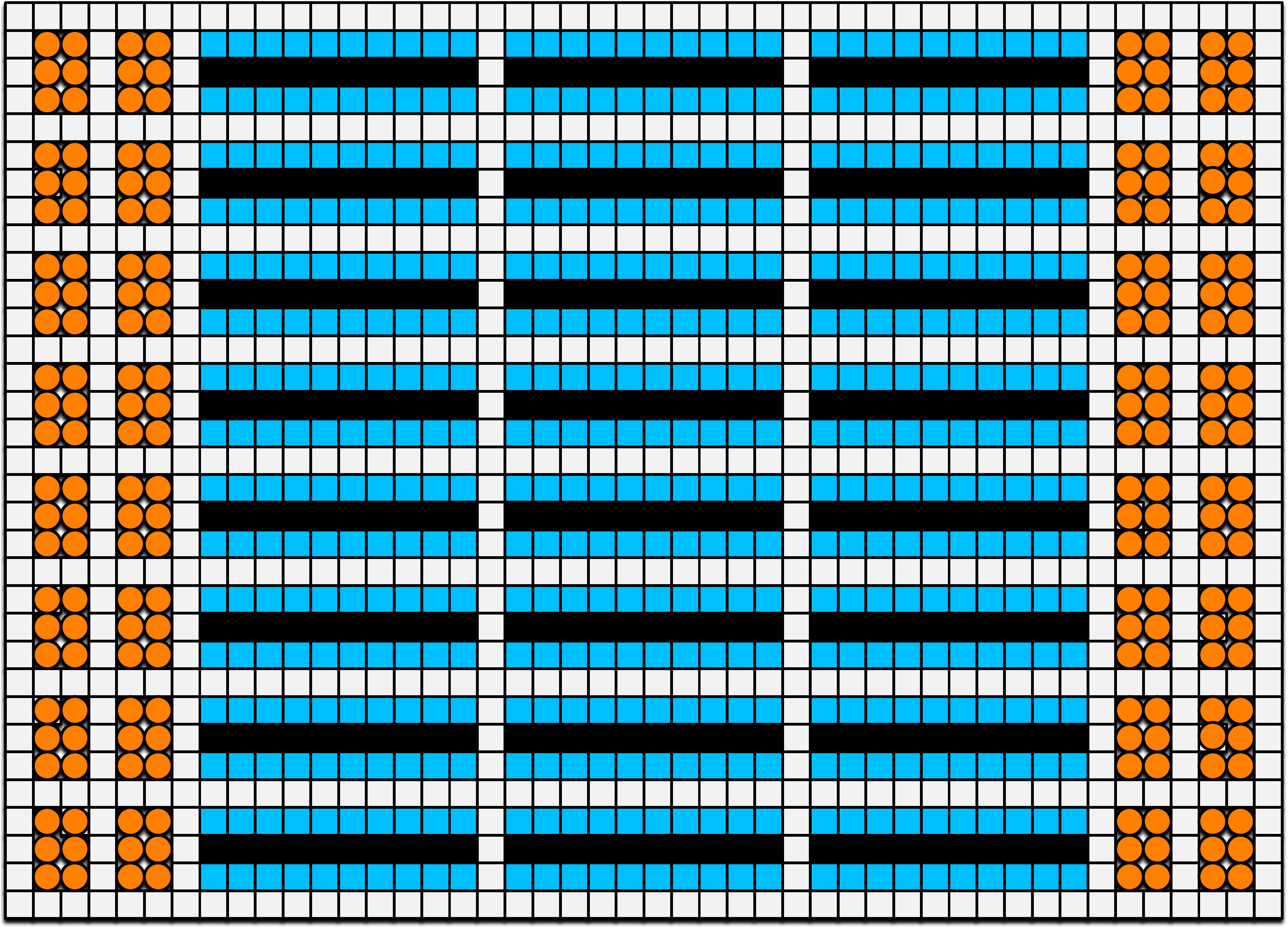} 
        \caption{Fulfillment warehouse map.}\label{fig:map1}
    \end{subfigure}
    \begin{subfigure}{\columnwidth}
        \centering
        \includegraphics[height=3.5cm]{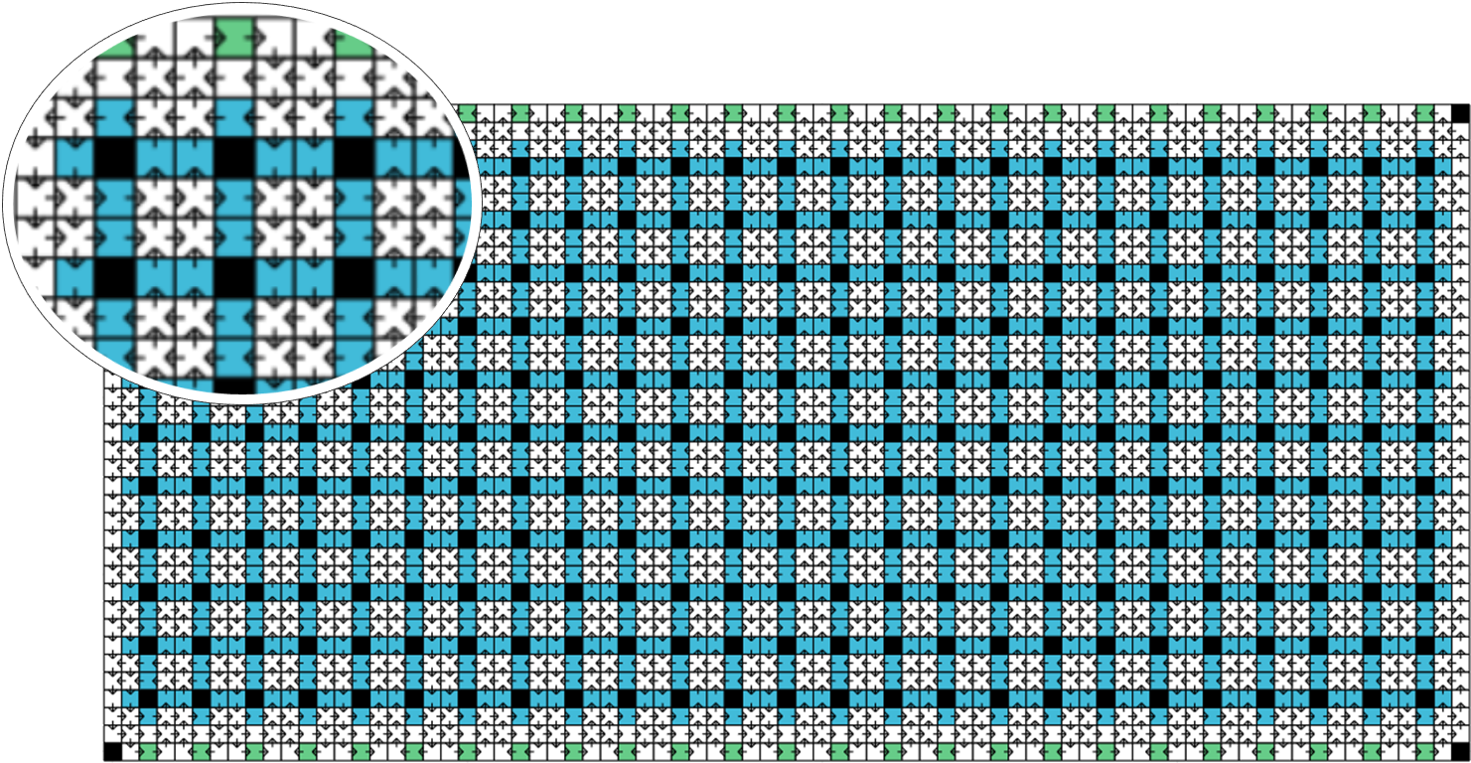} \caption{Sorting center map.}\label{fig:map2}
    \end{subfigure}
    \caption{Two typical warehouse maps. Black cells represent obstacles, which the agents cannot occupy. Cells of other colors represent empty locations, which the agents can occupy and traverse. }
\end{figure}
%%%%%%%%%%%%%%%%%%%%%%%%%%%%%%%%%%%%%%%%%%%%%%%%%%%%%%%%%%%%%%%
\section{Empirical Results} \label{sec:exp}

We implement \windowed{} in C++ with four Windowed MAPF solvers based on CBS, ECBS, CA* and PBS.\footnote{The code is available at \url{https://github.com/Jiaoyang-Li/RHCR}.}
We use SIPP~\cite{PhillipsICRA11}, an advanced variant of location-time A*, as the low-level solver for CA* and PBS.
We use Soft Conflict SIPP (SCIPP)~\cite{CohenSoCS19}, a recent variant of SIPP that generally breaks ties in favor of paths with lower numbers of collisions, for CBS and ECBS.
We use CA* with random restarts where we repeatedly restart CA* with a new random priority ordering until it finds a solution. % or reaches the runtime limit of 1 minute.
We also implement two existing realizations of Method (3) for comparison, namely holding endpoints~\cite{MaAAMAS17} and reserving dummy paths~\cite{LiuAAMAS19}. 
We do not compare against Method (1) since it does not work in our online setting.
We do not compare against Method (2) since we choose dense environments 
to stress test various methods and its performance in dense environments is similar to that of \windowed{} with an infinite time horizon.
We simulate 5,000 timesteps for each experiment with potential function threshold $p=1$.
We conduct all experiments on Amazon EC2 instances of type ``m4.xlarge'' with 16 GB memory.

%%%%%%%%%%%%%%%%%%%%%%%%%%%%%%%%%%%%%%%%%%%%%%%%%%%%%%%%%%%%%%%%%%%%%%%%%
\subsection{Fulfillment Warehouse Application} \label{sec:kiva}

\begin{table}[t]
    \small
    \centering
    %\resizebox{\columnwidth}{!}{
    \begin{tabular}{c|cccc}
    \toprule
    Framework & $m=60$ & $m=100$ & $m=140$ \\ %& $m=180$ \\
    \midrule
    \windowed{}         & 2.33 & 3.56 & 4.55 \\ %& 5.3797 \\
    HE     & 2.17 (-6.80\%) & 3.33 (-6.33\%) & 4.35 (-4.25\%) \\ %& 5.2489 \\
    RDP & 2.19 (-6.00\%) & 3.41 (-4.16\%) & 4.50 (-1.06\%)\\ %& 5.4701 \\
    \midrule
    \windowed{}         & $0.33\pm0.01$ & $2.04\pm0.04$ & $7.78\pm0.14$ \\ % & $21.18\pm0.34$ \\
    HE     & $0.01\pm0.00$ & $0.02\pm0.00$ & $0.04\pm0.01$ \\ % & $0.06\pm0.02$ \\
    RDP & $0.02\pm0.00$ & $0.05\pm0.01$ & $0.17\pm0.05$ \\ % & $0.3\pm0.12$ \\
    \bottomrule
    \end{tabular}%}
    \caption{Average throughput (Rows 2-4) and average runtime (in seconds) per run (Rows 5-7) of \windowed{}, holding endpoints (denoted by HE) and reserving dummy paths (denoted by RDP). % of different lifelong MAPF methods with different numbers of agents $m$.
    Numbers in parenthesis characterize throughput differences (in percentage) compared to \windowed{}.
    Numbers after ``$\pm$'' indicate standard deviations.}
    \label{tab:compare}
\end{table}

\begin{table*}[t]
    \centering
    \small
    \begin{tabular}{c|c|ccccccc}
    \toprule
    &$w$ & $m=400$ & $m=500$ & $m=600$ & $m=700$ & $m=800$ & $m=900$ & $m=1000$ \\
    \midrule
    \multirow{4}{*}{\rotatebox{90}{Throughput}} & 5        & 12.27 (-1.56\%) & 15.17 (-1.84\%) & 17.97 (-2.35\%) & 20.69 (-2.85\%) & 23.36 & 25.79 & 27.95 \\
    &10       & 12.41 (-0.41\%) & 15.43 (-0.19\%) & 18.38 (-0.11\%) & 21.19 (-0.52\%) & 23.94 & 26.44 & 28.77 \\
    &20       & 12.45 (-0.07\%) & 15.48 (+0.12\%) & 18.38 (-0.11\%) & 21.24 (-0.26\%) & 23.91  & - & - \\
    &$\infty$ & 12.46 & 15.46 & 18.40 & 21.30 & - & - & - \\
    \midrule
    \multirow{4}{*}{\rotatebox{90}{Runtime}} &5        & $0.61\pm0.00$ & $1.12\pm0.01$ & $1.87\pm0.01$ & $3.01\pm0.01$ & $4.73\pm0.02$ & $7.30\pm0.04$ & $10.97\pm0.06$ \\
    &10       & $0.89\pm0.00$ & $1.66\pm0.01$ & $2.91\pm0.01$ & $4.81\pm0.02$ & $7.79\pm0.04$ & $12.66\pm0.07$ & $21.31\pm0.14$ \\
    &20       & $1.36\pm0.01$ & $2.71\pm0.01$ & $5.11\pm0.03$ & $9.28\pm0.06$ & $17.46\pm0.14$ & - & - \\
    &$\infty$ & $1.83\pm0.01$ & $3.84\pm0.03$ & $7.63\pm0.06$ & $16.16\pm0.17$ & - & - & - \\
    \bottomrule
    \end{tabular}
    \caption{Average throughput and average runtime (in seconds) per run of \windowed{} using PBS.
    ``-'' indicates that it takes more than 1 minute for the Windowed MAPF solver to find a solution in any run.
    Numbers in parenthesis characterize throughput differences (in  percentage) compared to time horizon $w = \infty$.
    Numbers after ``$\pm$'' indicate standard deviations.}
    \label{tab:pbs}
\end{table*}

\begin{table*}[ht]
    \centering
    \small
    \begin{subtable}[t]{\textwidth}
        \centering
        \begin{tabular}{c|c|cccccc}
        \toprule
        %\multicolumn{6}{c}{ECBS} \\
        & $w$ & $m=100$ & $m=200$ & $m=300$ & $m=400$ & $m=500$ & $m=600$ \\
        \midrule
        \multirow{2}{*}{Throughput} & 5        & 3.19 (+1.02\%) & 6.23 (-1.21\%) & 9.17 (-1.47\%) & 12.03 (-2.03\%) & 14.79 (-2.68\%) & 17.28 \\
        & $\infty$ & 3.16 & 6.31 & 9.31 & 12.28 & 15.20 & - \\
        \midrule
        \multirow{2}{*}{Runtime} & 5        & $0.07\pm0.00$ & $0.26\pm0.00$ & $0.64\pm0.00$ & $1.27\pm0.01$ & $2.37\pm0.02$ & $4.22\pm0.10$ \\
        & $\infty$ & $0.38\pm0.00$ & $1.81\pm0.01$ & $5.09\pm0.03$ & $11.48\pm0.09$ & $23.47\pm0.22$ & - \\
        \bottomrule
        \end{tabular}
        \caption{\windowed{} using ECBS.}
    \end{subtable}\\
    \begin{subtable}[t]{0.62\textwidth}
        \centering
        \begin{tabular}{c|c|cccc}
        \toprule
        %\multicolumn{4}{c||}{CA*} & \multicolumn{3}{c}{CBS} \\
        & $w$ & $m=100$ & $m=200$ & $m=300$ & $m=400$ \\
        \midrule
        \multirow{2}{*}{Throughput} & 5        & 3.19 (+0.53\%) & 6.17 (-0.48\%) & 9.12 (-0.35\%) & - \\
        & $\infty$ & 3.17 & 6.20 & 9.16 & - \\
        \midrule
        \multirow{2}{*}{Runtime} & 5        & $0.05\pm0.00$ & $0.21\pm0.01$ & $1.07\pm0.10$ & - \\
        & $\infty$ & $0.19\pm0.00$ & $0.84\pm0.02$ & $2.58\pm0.12$ & - \\
        \bottomrule
        \end{tabular}
        \caption{\windowed{} using CA*.}
    \end{subtable}
    \begin{subtable}[t]{0.37\textwidth}
        \centering
        \begin{tabular}{c|c|cc}
        \toprule
        & $w$ & $m=100$ & $m=200$ \\
        \midrule
        \multirow{2}{*}{Throughput} & 5        & 3.17 & - \\
        & $\infty$ & - & - \\
        \midrule
        \multirow{2}{*}{Runtime} & 5 & $0.14\pm0.03$ & - \\
        & $\infty$ & - & - \\
        \bottomrule
        \end{tabular}%}
        \caption{\windowed{} using CBS.}
    \end{subtable}
\caption{Results of \windowed{} using ECBS, CA*, and CBS. Numbers are reported in the same format as in \Cref{tab:pbs}.
    }
    \label{tab:others}
\end{table*}

In this subsection, we introduce fulfillment warehouse problems, that are
commonplace in automated warehouses and are characterized by blocks of inventory pods in the center of the map and work stations on its perimeter. Method (3) is applicable in such well-formed infrastructures, and we thus compare \windowed{} with both realizations of Method (3).
We use the map in \Cref{fig:map1} from~\cite{LiuAAMAS19}.
It is a $33\times46$ 4-neighbor grid with 16\% obstacles.
The initial locations of agents are uniformly chosen at random from the orange cells, and the task assigner chooses the goal locations for agents uniformly at random from the blue cells.
For \windowed{}, we use time horizon $w=20$ and replanning period $h=5$. For the other two methods, we replan at every timestep, as required by Method (3).
All methods use PBS as their (Windowed) MAPF solvers.

\Cref{tab:compare} reports the throughput and runtime of these methods with different numbers of agents $m$.
In terms of throughput, \windowed{} outperforms the reserving dummy path method, which in turn outperforms the holding endpoints method.
This is because, as discussed in~\Cref{sec:prior}, Method (3) usually generates unnecessary longer paths in its solutions.
In terms of runtime, however, our method is slower per run (i.e., per call to the (Windowed) MAPF solver) because the competing methods usually replan for fewer than 5 agents. The disadvantages of these methods are that they need to replan at every timestep, achieve a lower throughput, and are not applicable to all maps.

%%%%%%%%%%%%%%%%%%%%%%%%%%%%%%%%%%%%%%%%%%%%%%%%%%%%%%%%%%%%%%%%%%%%%%%%%
\subsection{Sorting Center Application} \label{sec:sorting}
In this subsection, we introduce sorting center problems, that are also commonplace in warehouses and are characterized by uniformly placed chutes in the center of the map and work stations on its perimeter.
Method (3) is not applicable since they are typically not well-formed infrastructures. 
We use the map in \Cref{fig:map2}. It is a $37\times77$ 4-neighbor grid with 10\% obstacles. The 50 green cells on the top and bottom boundaries represent work stations where humans put packages on the drive units. The 275 black cells (except for the four corner cells)
represent the chutes where drive units occupy one of the adjacent blue cells and drop their packages down the chutes.
The drive units are assigned to green cells and blue cells alternately. In our simulation, the task assigner chooses blue cells uniformly at random and chooses green cells that are closest to the current locations of the drive units.
The initial locations of the drive units are uniformly chosen at random from the empty cells (i.e., cells that are not black).
We use a directed version of this map to make MAPF solvers more efficient since they do not have to resolve swapping conflicts, which allows us to focus on the efficiency of the overall framework.
Our handcrafted horizontal directions include two rows with movement from left to right alternating with two rows with movement from right to left, and our handcrafted vertical directions include two columns with movement from top to bottom alternating with two columns with movement from bottom to top.  We use replanning period $h=5$.

\Cref{tab:pbs,tab:others} report the throughput and runtime of \windowed{} using PBS, ECBS with suboptimality factor 1.1, CA*, and CBS for different values of time horizon $w$. As expected, $w$ does not substantially affect the throughput. In most cases, small values of $w$ change the throughput by less than 1\% compared to $w=\infty$. However, $w$ substantially affects the runtime. 
In all cases, small values of $w$ speed up \windowed{} by up to a factor of 6 without compromising the throughput. Small values of $w$ also yield scalability with respect to the number of agents, as indicated in both tables by missing ``-''.
For example, PBS with $w=\infty$ can only solve instances up to 700 agents, while PBS with $w=5$ can solve instances up to at least 1,000 agents.

%%%%%%%%%%%%%%%%%%%%%%%%%%%%%%%%%%%%%%%%%%%%%%%%%%%%%%%%%%%%%%%%%%%%%%%%
\subsection{Dynamic Bounded Horizons}
We evaluate whether we can use the deadlock avoidance mechanism to decide the value of $w$ for each Windowed MAPF episode automatically by using a larger value of $p$ and starting with a smaller value of $w$. We use \windowed{} with $w=5$ and $p=60$ on the instances in \Cref{sec:kiva} with 60 agents. We use ECBS with suboptimality factor 1.5 as the Windowed MAPF solver. The average time horizon that is actually used in each Windowed MAPF episode is 9.97 timesteps. The throughput and runtime are 2.10 and 0.35s, respectively. However, if we use a fixed $w$ (i.e., $p=0$), we achieve a throughput of 1.72 and a runtime of 0.07s for time horizon $w=5$ and a throughput of 2.02 and a runtime of 0.17s for time horizon $w=10$. Therefore, this dynamic bounded-horizon method is able to find a good horizon length that produces high throughput but induces runtime overhead as it needs to increase the time horizon repeatedly.

\section{Conclusions} \label{sec:conclusion} % and Future Work

In this paper, we proposed Rolling-Horizon Collision Resolution (\windowed{}) for solving lifelong MAPF by decomposing it into a sequence of Windowed MAPF instances. We showed how to transform several regular MAPF solvers to Windowed MAPF solvers. Although \windowed{} does not guarantee completeness or optimality, we empirically demonstrated its success on fulfillment warehouse maps and sorting center maps.
We demonstrated its scalability up to 1,000 agents while also producing solutions of high throughput.
Compared to Method (3), \windowed{} not only applies to general graphs but also yields better throughput.
Overall, \windowed{} applies to general graphs, invokes replanning at a user-specified frequency, and is able to generate pliable plans that cannot only adapt to continually arriving new goal locations but also avoids wasting computational effort in anticipating a distant future.

\windowed{} is simple, flexible, and powerful. It introduces a new direction for solving lifelong MAPF problems. There are many avenues of future work: 
\begin{enumerate}
    \item adjusting the time horizon $w$ automatically based on the congestion and the planning time budget, % 
    \item grouping the agents and planning in parallel, and
    \item deploying incremental search techniques to reuse search effort from previous searches.
\end{enumerate}

\section*{Acknowledgments}
The research at the University of Southern California was supported by the National Science Foundation (NSF) under grant numbers 1409987, 1724392, 1817189, 1837779, and 1935712 as well as a gift from Amazon. Part of the research was completed during Jiaoyang Li's internship at Amazon Robotics. The views and conclusions contained in this document are those of the authors and should not be interpreted as representing the official policies, either expressed or implied, of the sponsoring organizations, agencies, or the U.S. government.

\bibliography{ref}
\end{document}